%% file: main.tex
\documentclass[sigconf]{acmart}

\usepackage{subfigure}
\usepackage{multirow}
\usepackage[normalem]{ulem}
\useunder{\uline}{\ul}{}
\input{math_commands.tex}

\AtBeginDocument{%
  \providecommand\BibTeX{{%
    \normalfont B\kern-0.5em{\scshape i\kern-0.25em b}\kern-0.8em\TeX}}}
\usepackage[ruled,linesnumbered]{algorithm2e}

\usepackage{xspace}
\usepackage{threeparttable}
\usepackage{enumitem}
\usepackage{makecell}
\usepackage{bbding}
\usepackage{caption}

\newtheorem{definition}{Definition}

\newcommand{\vpara}[1]{\vspace{0.04in}\noindent\textbf{#1}\xspace}

\usepackage{threeparttable}
\usepackage{multirow}
\usepackage{colortbl}
\definecolor{mygray}{gray}{0.85}

\newcommand{\ie}{i.e.}

\newcommand{\model}{$\text{GA}^{2}\text{E}$\xspace}

\setcopyright{acmlicensed}
\copyrightyear{2024}
\acmYear{2024}
\acmDOI{XXXXXXX.XXXXXXX}

\acmConference[KDD '24]{Make sure to enter the correct
  conference title from your rights confirmation emai}{August 25--29,
  2024}{Barcelona, Spain}
\acmISBN{978-1-4503-XXXX-X/18/06}

\begin{document}

\title{Exploring Task Unification in Graph Representation Learning via Generative Approach}

\author{Yulan Hu}
\email{huyulan@ruc.edu.cn}
\affiliation{
  \institution{Renmin University of China, Kuaishou Technology}
   \country{}
}

\author{Sheng Ouyang}
\email{ouyangsheng@ruc.edu.cn}
\affiliation{
  \institution{Renmin University of China}
   \country{}
}

\author{Zhirui Yang}
\email{yangzhirui@ruc.edu.cn}
\affiliation{%
  \institution{Renmin University of China}
  \country{}
  }

\author{Ge Chen}
\email{chenge221@mails.ucas.ac.cn}
\affiliation{%
  \institution{University of Chinese Academy of Sciences}
  \country{}
  }

\author{Junchen Wan}
\email{wanjunchen@kuaishou.com}
\affiliation{%
  \institution{Kuaishou Technology}  
  \country{}
  }


\author{Xiao Wang}
\email{xiao_wang@buaa.edu.cn}
\affiliation{%
  \institution{Beihang University}  
  \country{}
  }

\author{Yong Liu}
\email{liuyonggsai@ruc.edu.cn}
\affiliation{%
  \institution{Renmin University of China}
  \country{}
  } 

\renewcommand{\shortauthors}{Yulan Hu, et al.}
\begin{abstract}
Graphs are ubiquitous in real-world scenarios and encompass a diverse range of tasks, from node-, edge-, and graph-level tasks to transfer learning. However, designing specific tasks for each type of graph data is often costly and lacks generalizability. Recent endeavors under the "Pre-training + Fine-tuning" or "Pre-training + Prompt" paradigms aim to design a unified framework capable of generalizing across multiple graph tasks. Among these, graph autoencoders (GAEs), generative self-supervised models, have demonstrated their potential in effectively addressing various graph tasks. Nevertheless, these methods typically employ multi-stage training and require adaptive designs, which on one hand make it difficult to be seamlessly applied to diverse graph tasks and on the other hand overlook the negative impact caused by discrepancies in task objectives between the different stages. To address these challenges, we propose \model, a unified adversarially masked autoencoder capable of addressing the above challenges seamlessly. Specifically, \model proposes to use the subgraph as the meta-structure, which remains consistent across all graph tasks (ranging from node-, edge-, and graph-level to transfer learning) and all stages (both during training and inference). Further, \model operates in a \textbf{"Generate then Discriminate"} manner. It leverages the masked GAE to reconstruct the input subgraph whilst treating it as a generator to compel the reconstructed graphs resemble the input subgraph. Furthermore, \model introduces an auxiliary discriminator to discern the authenticity between the reconstructed (generated) subgraph and the input subgraph, thus ensuring the robustness of the graph representation through adversarial training mechanisms. We validate \model's capabilities through extensive experiments on 21 datasets across four types of graph tasks. \model consistently performs well across the various tasks, demonstrating its potential as an extensively applicable model for diverse graph tasks. 
\end{abstract}

\ccsdesc[300]{Computing methodologies~Knowledge representation and reasoning}
\ccsdesc[300]{Networks~Network algorithms}

\keywords{Graph Neural Networks, Unified Graph Learning, Generative Graph Learning}
 
\maketitle

\section{Introduction}~\label{intro}
Graph data are ubiquitous in the real world, serving as a means to model structured and relational information~\cite{cai2005mining,jin2021application}. Graph-structured data exhibits variations in terms of domain and task type, with various graphs present in different domains~\cite{wu2020graph, wu2022graph}, encompassing a wide array of tasks, including node-, edge-, and graph-level learning, as well as transfer learning, etc. The diverse characteristics of different graph tasks pose challenge in developing a unified approach to graph learning. 

A few attempts have been made to develop a unified method for various graph tasks. The recent \textbf{"Pre-training + Prompt"} approach~\cite{sun2022gppt, wu2023survey, prog2023, liu2023graphprompt}, inspired by the success of prompt-based solutions in Natural Language Processing (NLP)~\cite{liu2023pre, wei2022chain}, is emerging as a viable option to address this challenge. This paradigm typically employs prompt techniques to coordinate pre-training and downstream tasks, emphasizing a unified structure~\cite{sun2022gppt, prog2023, allinone}. However, this approach relies heavily on prompt design. While it may effectively align one or two graph tasks, unifying additional tasks necessitates increasingly intricate and precise prompts, which raises questions about the scalability and robustness of these methods in accommodating the multitude of graph tasks. Prior to the "Pre-training + Prompt" approach, the \textbf{"Pre-training + Fine-tuning"} paradigm attracted numerous in-depth studies~\cite{xia2022survey, hu2020gpt, qiu2020gcc, xia2022towards, GraphMAE, S2GAE} aiming to address various downstream tasks. This paradigm typically involves designing handcrafted pretext tasks to extract graph knowledge into a pre-trained model, which is then transferred to multiple downstream tasks. In this context, generative self-supervised learning (SSL), represented by graph autoencoders (GAEs), has emerged as a promising attempt~\cite{sslsurvey}. Unlike traditional supervised graph learning, which relies on supervision signals, GAEs procure these signals from the graph itself. Specifically, GAEs design an auxiliary self-supervised task to extract attribute or structural (or both) knowledge from a graph~\cite{GraphMAE, S2GAE, graphmae2}, which is then applied to a range of downstream tasks using the learned graph representation. Despite the initial successes of GAEs, certain limitations persist. The main issue lies in the inconsistencies between the objectives of the self-supervised task and the downstream tasks. The existing GAEs within this paradigm generally design a specific type of self-supervised task, such as node-level feature reconstruction or link-level prediction, and then apply the learned graph representation to the downstream graph-level task~\cite{GraphMAE, S2GAE}. This approach is sub-optimal due to discrepancies between the reconstruction and downstream tasks. \citet{when_to_pretrain} pointed out the existence of the "negative transfer" phenomenon, attributed to the distribution shift between the pre-training and downstream data. Furthermore, the quest for an optimal combination of pre-training and fine-tuning tasks is costly, as demonstrated in Figure~\ref{fig:pretrain}. While the "negative transfer" between the SSL task and the downstream task remains unexplored, we postulate that the discrepancy between the objectives of SSL tasks and downstream tasks may lead to sub-optimal performance.

To the best of our knowledge, to date, there has been no prior work capable of seamlessly unifying the pre-training tasks with disparate downstream tasks. This observation begs the question: \textit{could a feasible framework exist that can generalize the capabilities of the pre-trained model across various downstream tasks, while simultaneously safeguarding the performance across different types of downstream tasks?}

In this paper, we explore the inherent limitations in developing a framework that can smoothly generalize the capabilities of pre-trained models to distinct downstream tasks. One roadblock lies in the conflict between the objectives of pre-training tasks and downstream tasks. Existing GAEs rely solely on node-level tasks (e.g., feature reconstruction) or edge-level tasks (e.g., missing edge prediction) for pre-training and then fine-tune the learned graph representation for node-, edge-, and graph-level tasks~\cite{GraphMAE, S2GAE}. Considering a GAE that leverages a node-level SSL pretext task to guide the downstream graph classification task, the pretext task primarily focuses on the fine-grained node-level features, while the downstream graph classification requires graph-level features to determine its category. Such a contradiction is detrimental to overall performance. Another obstacle lies in the lack of robustness of the learned graph representation. On one hand, existing GAE pre-training tasks predominantly rely on reconstruction criteria, such as mean square error (MSE) or scaled cosine error (SCE), to restore the input graph characteristics. However, few studies~\cite{ARGA, graphmae2} reveal that GAEs may be susceptible to feature or structure perturbations. On the other hand, the universality of a representation implies robustness which enables it to be applied in various task scenarios. Unfortunately, although existing GAEs have been developed to tackle different graph tasks, the aforementioned limitations have yet to be effectively addressed.

\vpara{Contributions.} In light of the aformentioned observations, we propose a robust generative model, termed the Adversarial Graph AutoEncoder (\model), that is adaptive to multiple graph tasks. Unlike existing generative methods, \model can seamlessly unify numerous downstream tasks in an all-in-one fashion and is capable of circumventing the "negative transfer" phenomenon by reformulating the input objective. Furthermore, \model notably acknowledges the lack of rigor existing GAE methods demonstrate when applied to various graph tasks, and subsequently introduces an innovative adversarial training mechanism to reinforce the robustness of semantic features. In Table~\ref{intro:methods_comparision}, we investigate the key features of existing GAE-based self-supervised methods. Among these, \model benefits from several critical designs, the roles of which are further analyzed in Section~\ref{ablation_study}.

\textbf{Meta-Structure reformulation.} The primary discrepancy between pre-training and fine-tuning tasks lies in their objectives. Taking into account various graph tasks ranging from node, edge, and graph tasks, to even transfer learning tasks, \model introduces the subgraph as the meta-structure. Each raw graph will be reformulated into multiple subgraphs to perform pre-training tasks. This not only ensures different graph tasks share the same meta-structure, but also eliminates the gap between pre-training and downstream tasks.

\textbf{Adversarial discrimination.} In addition to maintaining consistency across graph tasks, we introduce an adversarial training strategy to enhance the model's robustness. Specifically, we term the entire GAEs as the generator, drawing on the concept of Generative Adversarial Networks (GANs), and introduce another network to discriminate the authenticity of the reconstructed graph. Our empirical studies suggest that this adversarial strategy substantially benefits unified GAEs.

\textbf{Uniformity of the training pipelines.} We establish a unified training pipeline as "Generate then Discriminate." This intuitive yet non-trivial pipeline focuses on the complete exploration of generative capability. Crucially, this pipeline is task-agnostic, requiring no specific task-related design elements.
\begin{table}[]
\caption{Technical comparison between generative SSL methods. AE: autoencoder framework; Feature Recon.: Using feature for reconstruction. Perturb. Robust: robust to the perturbations. NT Free: free of "negative transfer". }
\label{intro:methods_comparision}
\resizebox{\linewidth}{!}{
\begin{tabular}{@{}ccccccc@{}}
\toprule
Methods               & AE                        & \begin{tabular}[c]{@{}c@{}}Feature\\ Recon.\end{tabular} & \begin{tabular}[c]{@{}c@{}}Cross\\ Task\end{tabular} & \begin{tabular}[c]{@{}c@{}}Cross\\ Domain\end{tabular} & \begin{tabular}[c]{@{}c@{}}Perturb.\\ Robust\end{tabular} & \begin{tabular}[c]{@{}c@{}}NT\\ Free\end{tabular} \\ \midrule
VGAE~\cite{VGAE}                  & \Checkmark & -                                                        & -                                                    & -                                                      & -                                                         & -                                                 \\
ARVGA~\cite{ARGA}                 & \Checkmark & -                                                        & -                                                    & -                                                      & \Checkmark                                 & -                                                 \\
GALA~\cite{gala}                  & \Checkmark & \Checkmark                                & -                                                    & -                                                      & -                                                         & -                                                 \\
MaskGAE~\cite{maskgae}               & \Checkmark & \Checkmark                                & -                                                    & -                                                      & -                                                         & -                                                 \\
GPT-GNN~\cite{hu2020gpt}               & -                         & -                                                        & -                                                    & -                                                      & -                                                         & -                                                 \\
GraphMAE~\cite{GraphMAE}              & \Checkmark & \Checkmark                                & \Checkmark                            & \Checkmark                              & -                                                         & -                                                 \\
GraphMAE2~\cite{graphmae2}             & \Checkmark & \Checkmark                                & -                                                    & -                                                      & \Checkmark                                 & -                                                 \\
S2GAE~\cite{S2GAE}                 & \Checkmark & -                                                        & \Checkmark                            & \Checkmark                              & -                                                         & -                                                 \\ \midrule
\model & \Checkmark & \Checkmark & \Checkmark                              & \Checkmark                                 & \Checkmark      & \Checkmark       \\ \bottomrule
\end{tabular}}
\end{table} 

\section{RELATED WORKS}
\vpara{Generative Graph Learning.}
 Generative learning for graphs aims to generate new graphs using input data as supervisory signals~\cite{sslsurvey}. GAE/VGAE~\cite{VGAE} represent pioneering generative models leveraging graph auto-encoders to recover the adjacency matrix from the input graph. Subsequently, GAE-based models have proliferated in the field of graph learning. ARGA/ARVGA~\cite{ARGA} integrates regularization into the GAE/VGAE framework, effectively constraining the latent space to conform to the prior distribution observed in real data. To mitigate the inflexibility caused by the Gaussian distribution assumption in variational inference, SIG-VAE~\cite{SIGVE} introduces semi-implicit variational inference (SIVI) into VGAE to accommodate a more flexible posterior distribution. In the most recent years, inspired by the success of masked auto-encoders (MAEs) in computer vision~\cite{mae}, GraphMAE~\cite{GraphMAE} has achieved promising performance by employing masking strategies on node attributes. However, GraphMAE has proven to be susceptible to feature perturbations. To address this issue, GraphMAE2~\cite{graphmae2} introduces a multi-view random re-mask strategy in the decoding stage, thereby learning a more robust graph representation. SeeGera~\cite{SeeGera} builds on the capabilities of SIG-VAE~\cite{SIGVE} by incorporating a masking mechanism that effectively captures rich semantic information from node features. This enhanced approach is simultaneously applied to both link prediction and node classification tasks. HGMAE~\cite{HGMAE} introduces several masking techniques and training strategies for heterogeneous graphs, extending MAE to heterogeneous graph learning. MaskGAE~\cite{MaskMAE} proposes a path-wise masking strategy and provides a theoretical understanding of masking. S2GAE~\cite{S2GAE}, a MAE variant, incorporates a direction-aware graph masking strategy and a tailored cross-correlation decoder, yielding impressive performance across a wide range of graph tasks.

Another related generative work is Generative Adversarial Networks (GANs)~\cite{gan}. GANs typically incorporate a generator and a discriminator to optimize in an adversarial manner, a concept that is beneficial for promoting graph SSL. GraphGAN~\cite{GraphGAN} introduces the adversarial concept to graph learning, employing a generator to learn the connectivity between neighboring nodes and the target node, while a discriminator is encouraged to differentiate the realness of the connectivity. In contrast to GraphGAN~\cite{GraphGAN}, AGE~\cite{AGE} achieves better performance by generating fake neighbors directly from a continuous distribution and considering asymmetric (directional) semantic information. NetGAN~\cite{NetGAN} utilizes a generator to produce a sequence and then adopts a discriminator to determine the authenticity probability of the random walk sequence, differing from GraphGAN~\cite{GraphGAN}, which focuses on the connectivity of neighboring nodes. Other works~\cite{ARGA,ANE} leverage the adversarial training scheme as a regularization technique, thus ensuring the robustness of latent representation. Additionally, a few GAN-related works have been applied in real scenarios. GCGAN~\cite{GCGAN} integrates GAN with Graph Convolutional Network (GCN) to predict traffic conditions.

\vpara{Unified Graph Learning.}
It is critical to highlight that, considering the multi-level nature (from node- and edge-level to graph-level) and variety of graph tasks (from classification and prediction to transfer learning), extremely limited efforts have been made to develop a "One-For-All" model suitable for graph learning. S2GAE~\cite{S2GAE} represents a unified work under the "Pre-training + Fine-tuning" paradigm. It develops a masked edge prediction pre-training task, further unifying different graph tasks with the learned representation. Additionally, the development of large-scale language models (LLMs) has inspired several studies~\cite{graphprompt, generalizedgraphprompt, allinone,oneforall} to adopt the "Pre-training + Prompting" paradigm to address multi-level graph learning. ~\citet{allinone} presents a novel multi-task prompting method that reformulates and unifies multi-level graph tasks, effectively bridging the gap between the pre-trained model and downstream tasks. OFA~\cite{oneforall} proposes a general framework using text-attributed graphs to unify different graphs, thereby generalizing its ability to multi-level graph tasks.

\section{PRELIMINARIES} 
\subsection{Necessary Background}
\vpara{Notations} We denote a graph as $\mathcal{G} = (\mathcal{V}, \mathcal{E}, \mathbf{X})$, where $\mathcal{V}$ represents the set of nodes on the graph. $\mathcal{E} \subseteq \mathcal{V} \times \mathcal{V}$ stands for the set of edges within the graph that connect the nodes. $\mathbf{X} \in \mathbb{R}^{N \times d_f}$ refers to the node feature matrix, where $N$ signifies the number of nodes, and $d_f$ denotes the dimension of the node features. Furthermore, we denote the adjacency matrix of the graph $\mathcal{G}$ as $\mathbf{A} \in \{0,1\}^{N \times N}$.

\vpara{Additional background}
Additional necessary background information, including a brief introduction to the Graph Neural Network (GNN), Masked Graph Autoencoder, and Generative Adversarial Network (GAN), can be found in Appendix ~\ref{app:background}.

\section{METHODOLOGY}
\begin{figure*}[h]
\centering 
\includegraphics[width=0.9\textwidth]{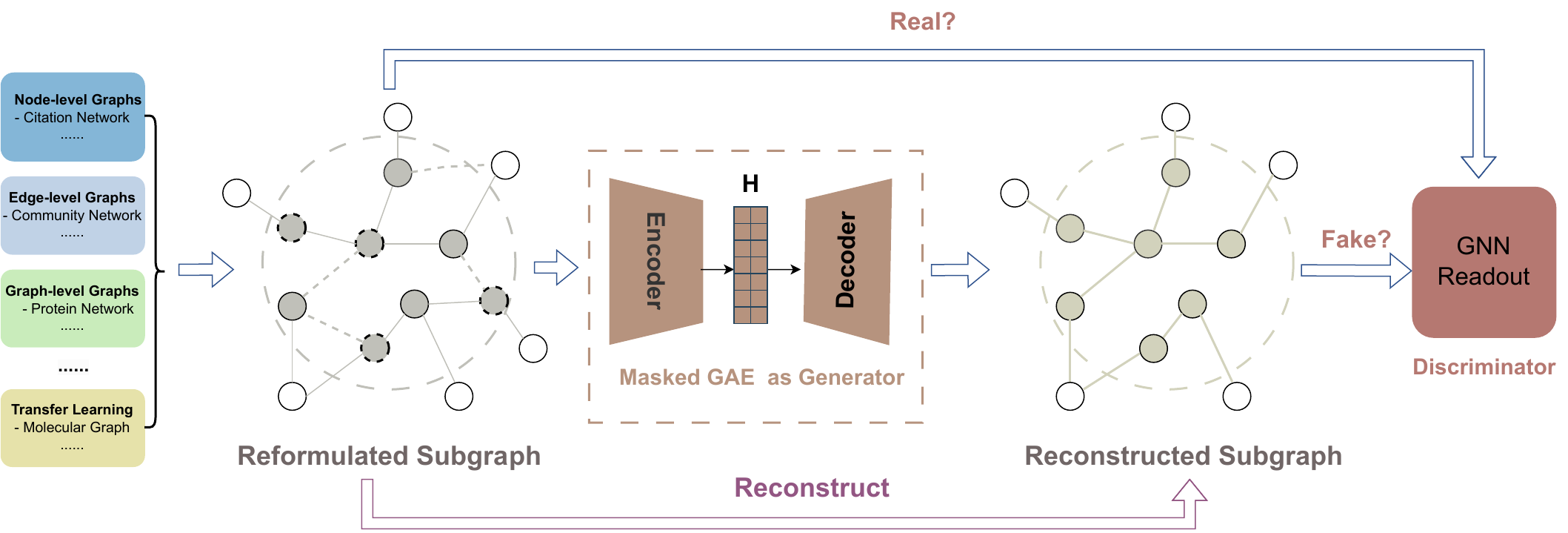}
\caption{The Overview of \model. \model accepts arbitrary graphs as input and reformulates them into subgraphs. Training is conducted adversarially within \model, where a masked GAE is adopted to reconstruct the input subgraph and generate the reconstructed graph. Additionally, a GNN Readout operates as a discriminator to determine the authenticity between the reformulated graph and the reconstructed graph.
} 
\label{fig:overview}
\end{figure*}
In this section, we formally introduce our method, \model. Figure~\ref{fig:overview} provides an overview of our method. \model exhibits an intuitive structure composed of two modules: a masked GAE as the generator and a GNN readout as the discriminator. In the following sections, we first describe the meta-structure reformulation for multiple tasks in Section~\ref{subgraph_formulation}, then elaborate on the learning process in Section~\ref{subgraph_learning}. 
Finally, we present a comprehensive explanation of why \model is capable of handling multiple graph tasks.

\subsection{Meta-structure Reformulation}~\label{subgraph_formulation}
Essentially, multi-task graph learning includes various levels of graph tasks as well as distinct graphs. To develop a versatile and unified generative model, the establishment of a consolidated meta-structure is the principle matter to address.

\vpara{Subgraph as meta-structure.}~\label{subgraph_learning}
\begin{figure}[H]
  \centering
  \subfigure[A subgraph instance.]{
		\label{method_subgraph_instance}
		\includegraphics[width=0.23\textwidth]{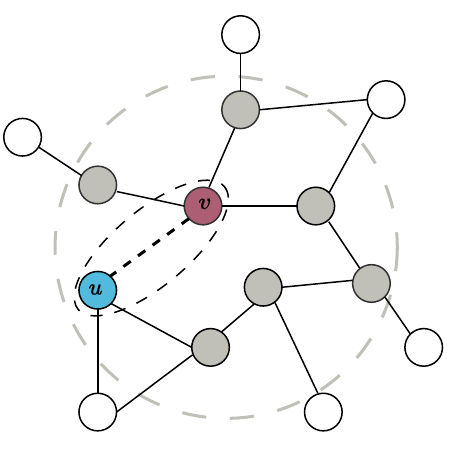}}
    \subfigure[Meta structures.]{
		\label{method_node_edge_graph}
		\includegraphics[width=0.17\textwidth]{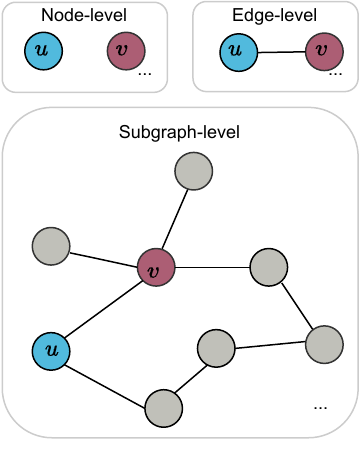}}		
    \caption{The subgraph represents the meta-structure that contains node, edge and graph level knowledge.} 
  \setlength{\abovecaptionskip}{0cm}
\end{figure}
To develop the meta-structure that encapsulates comprehensive graph knowledge for multi-task graphs, we first present the definition of meta-structure for multi-tasks.
\begin{definition}
    A meta-structure for multi-task graph learning should constitute the minimal union of learning objectives (nodes, edges, and graphs) for each sub-task and maintain consistency throughout both the pre-training, fine-tuning and inference stages.
\label{definition:1}
\end{definition}
Note that in Definition~\ref{definition:1}, we emphasize that the meta-structure remains consistent during all stages, which implies two aspects. First, the input meta-structure for multiple tasks should be identical. Second, the input meta-structure for different training stages --- \ie, pre-training, fine-tuning, and inference --- should remain consistent.

Figure~\ref{method_subgraph_instance} presents a graph instance that contains characteristics at different levels. Specifically, we list the node, edge, and graph instances in Figure~\ref{method_node_edge_graph}. Given nodes $u$, $v$, and the edge $e_{uv}$ between them, it is intuitive that the subgraph includes both nodes and edges, indicating that the subgraph represents the minimum superset of different graph instances. Based on this observation, we choose the subgraph as the meta-structure to represent both node-, edge-, and graph-level instances in different stages. Specifically, for node-level tasks, e.g., node classification, we randomly select $N_s$ nodes from the graph's node set $\mathcal{V}$ as the seed set $\mathcal{S}$, where $N_s < N$. For each node in $\mathcal{S}$, we construct a subgraph centered on the node, thereby obtaining $|\mathcal{S}|$ subgraphs. The rationale during fine-tuning and inference is similar, we construct the subgraph around each target node in the same way as in the pre-training stage, and the subsequent operation is performed on the meta-structure. On this basis, node classification is transformed into a graph-level problem, mitigating the structural discrepancy between the pre-training and fine-tuning phases.

Analogously, for edge-level tasks, we commence with edges. During the pre-training phase, we randomly select a portion of edges from the graph's edge set $\mathcal{E}$ as the seed edge set, $\mathcal{L}$. For each edge in $\mathcal{L}$, we construct a subgraph centered around that edge, eventually obtaining $\mathcal{L}$ edge-centered subgraphs upon which we can perform pre-training. 

Utilizing subgraphs as the meta-structure manages to meet the criteria outlined in Definition~\ref{definition:1}. This not only addresses the aforementioned challenges but also provides necessary local structural knowledge. Nevertheless, the method for constructing the meta-structure requires further investigation.

\vpara{Subgraph reformulation.}
Graph-level tasks accept the graph itself as input without additional design, while node- and edge-level tasks require techniques to transform the individual node or edge into a subgraph instance. Extensive efforts have been made to construct subgraphs~\cite{frasca2022understanding,alsentzer2020subgraph,huang2020graph}. In this paper, our method of construction is not fixed; we can use the $k$-hop ego graph as the subgraph or perform random walk for each target node to construct subgraphs. For illustration, we adopt the random walk with restart (RWR) method~\cite{random_walk_survey}. The detailed RWR algorithm is presented in Algorithm~\ref{alg:rwr}. However, the method to construct subgraphs that originate from a single node or edge has slight differences.

The construction of a node-level subgraph is illustrated in Figure~\ref{rwr_node}. We perform a random walk to iteratively add the current node's neighbors to the preset $\gS$, or return to the starting node with a probability $p$. This process continues until a fixed number of nodes are collected, denoted as $RWR(v)$. Subsequently, we extract the node set $\gS$ from $\mathcal{G}$, ensuring that each subgraph contains an equivalent number of nodes. The construction of an edge-level subgraph is presented in Figure~\ref{rwr_edge}. Subgraph construction at the edge-level can be regarded as a repetitive process of node-level subgraph construction. The edge $e_{v_{0}v_{4}}$, connecting $v_0$ and $v_4$, can be denoted as a positive edge. We initiate from $v_0$ and $v_4$ separately to perform RWR until the predefined size of the subgraph is reached.

\subsection{The Design of \model}~\label{subgraph_learning}
In this section, we deconstruct \model to reveal the underlying design rationale. \model derives its benefits primarily from two aspects: 1) The adoption of the subgraph as the meta-structure. 2)The adversarial training mechanism to enhance model robustness.    

Specifically, the first aspect emphasizes the conformation of graph data, which is introduced in Section~\ref{subgraph_formulation}. The second aspect involves the design of \model, particularly its adversarial attributes.

\vpara{Masked GAE as generator.} Existing generative methods typically adopt the masked GAE as the backbone model, emphasizing the reconstruction of the input graph feature~\cite{GraphMAE, graphmae2} or structure~\cite{S2GAE}. \model employs a masked GAE both to reconstruct the input subgraph attribute and to generate deceptive subgraphs. As such, the masked GAE within \model fulfills two distinct roles.

Let $f_E$ and $f_D$ denote the encoder and decoder, respectively. Both $f_E$ and $f_D$ typically include one or more GNN layers. The encoder, $f_E$, accepts a batch of subgraphs as input:
\begin{equation}
    \gG = Assemble([sg_0,sg_1,...,sg_B]),
\end{equation}
where $B$ represents the batch size. For the sake of illustration, we utilize $\gG$ to represent the assembled graphs. Inspired by the widespread applications in other fields~\cite{mae,bert} and the successful implementation of masked GAEs~\cite{GraphMAE}, we apply a random mask to the node features of $\gG$. Specifically, we randomly select a subset of nodes $\tilde{\gV}$ from $\gV$, the features of nodes in $\tilde{\gV}$ are masked and assigned a special "[MASK]" token, while the features of the remaining nodes are left unchanged. This results in an altered graph $\tilde{\gG}=(\tilde{A}, \tilde{X})$. Subsequently, $f_E$ accepts this altered graph as input and maps it into a latent representation from which the decoder, $f_D$, can further rebuild the input graph as follows:
\begin{equation}
\label{eq:gen}
   \mathbf{H}=f_E(\tilde{\mathbf{A}} ,\tilde{\mathbf{X} }  ), \mathbf{\bar{\mathcal{G}}  }=f_D(\tilde{\mathbf{A}},\mathbf{H}),
\end{equation}
where $\mathbf{H}$ denotes the hidden representation learned by $f_E$, while $\mathbf{\bar{\mathcal{G}}}$ refers to the graph reconstructed by $f_D$.

Till now, the encoding and decoding processes of \model have been introduced. As mentioned earlier, the masked GAE in \model assumes two crucial roles.

From the perspective of feature reconstruction, $\tilde{\gG}$ represents the reconstructed graph with $\gG$ acting as supervision signals. As we aim to restore the input graph attribute, we adopt the scaled cosine error (SCE)~\cite{GraphMAE} as the reconstruction criterion for reconstructing $\gG$ as:
\begin{equation}
\label{eq:reconstruct}
   \gL_{rec} = SCE(X, \tilde{X}),
\end{equation}
A detailed explanation of SCE can be found in Appendix~\ref{reconstruction_criterion}. This reconstruction process is strongly aligned with practices in NLP~\cite{bert} and CV~\cite{mae}, where the words in a sentence are masked and then predicted, or the missing pixels within an image are corrupted and subsequently predicted.
 
Furthermore, apart from the unification of the graph meta-structure, \model also assures its generality via an adversarial training mechanism. Following this, we perceive the entire GAE in \model as a generator from the perspective of robust learning, with $\tilde{\gG}$ viewed as the generated graph output by the generator, i.e., the masked GAE. The generator aims to produce an output akin to $\gG$—that is, to deceive the discriminator (which will be introduced later) into believing that the reconstructed output is as authentic as $\gG$.

\vpara{GNN readout as discriminator.} The key concept behind the adversarial mechanism involves the use of an auxiliary model to discriminate between the raw input graph and the reconstructed graph. This requires the formulation of a discrimination task and the design of the discriminator. We regard discrimination as a binary classification task and utilize a simple combination of GNN layers, a graph pooling layer, and a linear classification layer as the discriminator. 

The discriminator accepts both the reconstructed graphs and the original graph as input, aiming to discern their authenticity. We assign a label of 1 (representing a positive sample) to the original graph $\mathcal{G}$ and a label of 0 (indicating a negative sample) to the reconstructed graph $\tilde{\gG}$. By doing so, \model effectively refines the masked GAE at a semantic level.

This process catalyzes a feedback loop in which the discriminator's ability to discern authenticity gradually improves. Consequently, the generator is compelled to produce increasingly realistic graphs that resemble $\gG$, thereby enhancing the generator's capability. The loss function for the discriminator is formulated as: 
\begin{equation}
\label{eq:dis}
\mathcal{L}_{d}= {-\textstyle \sum_{\mathcal{G}\sim \mathcal{S}}^{} \log_{}{\mathcal{D} (\mathcal{G} )} } {-\textstyle \sum_{\mathcal{\tilde{G}}\sim \mathcal{\tilde{S}}}^{}\log (1-\mathcal{D}(\mathcal{\tilde{G}})) }   ,
\end{equation}
where $\mathcal{D}$ represents the discriminator model, $\mathcal{S}$ symbolizes the set of original graph samples, and $\mathcal{\tilde{S}}$ represents the set of reconstructed graph samples. Concurrently, the generator strives to generate high-quality graphs according to the following formula:
\begin{equation}
\label{eq:gen}
{\gL}_{gen}= -\textstyle \sum_{\mathcal{\tilde{G}}\sim \mathcal{\tilde{S}}}^{}\log (\mathcal{D}(\mathcal{\tilde{G}})).
\end{equation}
${\gL}_{rec}$ and ${\gL}_{gen}$ are simultaneously optimized as follows:
\begin{equation}
\label{eq:gen_loss}
    \mathcal{L}_{g}= {\textstyle \sum_{\mathcal{G} \sim \mathcal{S}  }^{} \mathcal{L}_{sce}(\mathcal{G},\tilde{\mathcal{G}} ) } +\alpha \mathcal{L}_{gen},
\end{equation}
Here, $\alpha$ is a coefficient used to modulate the contribution of the discriminator.
The training and inference processes of \model are summarized in Algorithm~\ref{alg:procedure}.

\begin{algorithm}[htb]
\caption{Training and inference process of \model.}\label{alg:procedure}
\KwIn{Subgraph set $\gG=\{sg_0,sg_1,...sg_B\}$, Training epochs $C$, Test subgraph set $\gT$;} 
\KwOut{Test result set $R$;}
$\triangleright$ \textcolor{blue}{Training Process} \\
\For{$epoch$ in $C$}{
        {Corrupt the input graph to form: $\tilde{\gG}$=MASK($\mathcal{G}$)}; \\
        {Obtain the reconstructed graph $\tilde{\mathcal{G}}$ using Equation~(\ref{eq:gen})}; \\
        {Calculate $\mathcal{L}_{sce}$ via Equation~(\ref{eq:sce})}; \\
        {Determine the Generator's loss using Equation~(\ref{eq:gen_loss})}; \\
        {Update the Generator}; \\                
        {Compute $\mathcal{L}_{d}$ utilizing Equation~(\ref{eq:dis})}; \\        
        {Update the Discriminator}; \\            
                
    }
    $\triangleright$ \textcolor{blue}{Inference Process}  \\
        \For{$sg \in \mathcal{T}$}{
        {$\mathbf{h}=f_E(sg)$} \\
        {$\mathbf{z}=Readout(\mathbf{h}$)} \\
        {$y=Softmax(\mathbf{z})$ } \\
        {$R.insert(y)$ } \\
        }
    {\Return $R$} \\ 
\end{algorithm}

\subsection{Why it Works ?}~\label{why_it_works}
\vpara{Unification of Graph Tasks.} One of the crucial reasons for the efficacy of our proposed method is its ability to unify various types of graph tasks into graph-level tasks, which simplifies the process and promotes task consistency. In the case of node classification tasks, we base the construction of induced subgraphs on nodes. Conducting graph classification on these induced subgraphs is equivalent to performing node classification. This approach encapsulates all relevant node information within a dedicated subgraph that serves as a distinguishable learning unit, thereby redefining the originally node-dependent task as a graph-level task. Similarly, for link prediction tasks, we construct induced subgraphs based on edges. Executing graph classification on these induced subgraphs is functionally akin to classifying on individual edges. This procedure represents each edge association as a distinct subgraph, effectively transforming edge-dependent tasks into graph-level tasks. By transforming varied types of graph tasks into uniformly graph-level tasks, we preserve task consistency, which is vital for a machine learning model’s capacity to learn effectively.

\vpara{Alignment between Pre-training and Downstream Tasks.} Another contributing factor is the alignment of the pre-training and downstream tasks, which effectively minimizes the gap between them, thereby reducing any potential negative transfer. By consolidating all graph-level tasks into a single tier, \model universally operates at this level during both pre-training and inference stages of the downstream tasks. This strategic alignment ensures that the nuances of each task are duly captured during pre-training and proficiently applied during downstream tasks. Furthermore, \model also demonstrates flexibility, adapting to downstream tasks of varying complexities. By maintaining identical task formulation between the pre-training methods and downstream tasks, \model can sufficaiently transfer the knowledge 
 from pre-trained model to downstream tasks. This approach significantly mitigates issues related to negative transfer, further enhancing the model’s performance across a variety of tasks.

In conclusion, through task unification at the graph level and alignment of pre-training with downstream tasks, \model offers an effective methodology to apply the learned knowledge. This process reduces discrepancies, minimizes negative transfer, and maximizes the model's efficacy across a broad array of tasks.

\section{EXPERIMENTS}
In this section, we conduct comprehensive experiments on the node-, edge-, graph-level tasks as well as transfer learning to address the following research questions: $\mathcal{RQ}1$: Can our method achieve consistent and excellent performance across various levels of downstream graph tasks?  $\mathcal{RQ}2$: How do the main components of the model impact its overall performance? $\mathcal{RQ}3$: How sensitive is our model to parameter changes? The detailed introduction of datasets in the experiment can be found in the Appendix~\ref{app:data}. The detailed introduction of compared baselines can be found in the Appendix~\ref{app:baselines}, and the detailed experimental settings is presented in Appendix~\ref{app:experient_settings}.
\subsection{Main Results ($\mathcal{RQ}1$)}
\vpara{Node Classification.}
\begin{table*}[]
\caption{Node classification on seven datasets. The best results are highlighted in bold, the second-best results are underlined.}
\label{tab:nc}
\begin{threeparttable}
\begin{tabular}{@{}l|cccccccc@{}}
\toprule
                            & Methods    & Cora                & Citeseer            & Pubmed              & Photo               & Computer            & CS                  & Physics             \\ \midrule
\multirow{2}{*}{Supervised} & GCN        & 80.84±0.44          & 70.46±0.85          & 79.02±0.26          & 90.79±2.47          & 85.34±1.64          & 92.10±0.16          & \underline{95.41±0.11}          \\
                            & GAT        & 83.00±0.70          & 72.50±0.70          & 79.00±0.30          & \underline{93.49±0.16}          & \underline{89.79±0.76}          & 91.48±0.17          & 95.25±0.14          \\ \midrule
\multirow{9}{*}{Self-supervised}  & GRACE      & 77.60±0.28          & 67.24±0.93          & 78.40±1.33          & 92.45±0.34          & 88.11±0.45          & 67.57±12.91         & 85.33±6.76          \\
                            & DGI        & 82.30±0.60          & 71.80±0.70          & 76.80±0.60          & 91.61±0.22          & 83.95±0.47          & 92.15±0.63          & 94.51±0.52          \\
                            & MVGRL      & 83.50±0.40          & 73.30±0.50          & 80.10±0.70          & 91.74±0.07          & 87.52±0.11          & 92.11±0.12          & 95.33±0.03          \\
                            & VGAE       & 71.50±0.40          & 65.80±0.40          & 72.10±0.50          & 92.20±0.11          & 86.37±0.21          & 92.11±0.09          & 75.35±0.14          \\
                            & CCA-SSG    & 84.00±0.40          & 73.10±0.30          & 81.00±0.40          & 93.14±0.14          & 88.74±0.28          & 93.31±0.22          & 95.38±0.06          \\
                            & GraphMAE   & 84.20±0.40          & 73.40±0.40          & 81.10±0.40          & 92.98±0.35          & 88.34±0.27          & 93.08±0.17          & 95.30±0.12          \\
                            & GraphMAE2  & \textbf{84.50±0.60} & \underline{73.40±0.30}          & \underline{81.40±0.50}          & -                   & -                   & -                   & -                   \\
                            & SeeGera    & \underline{84.30±0.40}          & 73.00±0.80          & 80.40±0.40          & 92.81±0.45          & 88.39±0.26          & \textbf{93.84±0.11} & 95.39±0.08          \\ \cmidrule(l){2-9} 
\multicolumn{1}{c|}{}       &\model & 83.95±0.40          & \textbf{73.60±0.15} & \textbf{81.50±0.60} & \textbf{93.60±0.22} & \textbf{89.87±0.11} & \underline{93.72±0.10}          & \textbf{95.61±0.00}          \\ \bottomrule
\end{tabular}
\begin{tablenotes}
        \footnotesize
        \item[] \scriptsize "-" indicates that the results not reported in the original paper. 
    \end{tablenotes}
\end{threeparttable}
\end{table*}
Table~\ref{tab:nc} presents the node classification results. We observe that even though \model is not specifically designed for node-level tasks, it still demonstrates robust performance on most of the datasets. We attribute this to two aspects. First, the unified meta-structure eliminates discrepancies between training and inference. Second, the adversarial training strategy enables \model to learn resilient graph representations. Additionally, by carefully refining the GAEs, generative-based models display more promising results than contrastive-based methods, demonstrating that the potential of GAEs.

\vpara{Link Prediction.} The results for link prediction are presented in Table~\ref{tab:lp}. As a whole, generative models tend to outperform contrastive methods in link prediction tasks, with the exception of GraphMAE. This superior performance can be attributed to the fact that generative models typically utilize graph reconstruction as their objective, which aligns well with the downstream task of link prediction. On the other hand, GraphMAE targets the reconstruction of node features as its objective. \model exhibit superior results compared to established baselines across all datasets. Interestingly, although our method shares the same objective as GraphMAE - reconstructing node features - it met or even exceeded the performance of methods focusing on graph structure reconstruction, such as SeeGera. The success of this approach lies in its design: rather than focusing only on individual nodes, our method learns from the local neighborhood structure of nodes via induced subgraphs, thus encapsulating superior neighbor structure information. This comprehensive focus allows for a deeper learning of node-neighborhood interactions, proving especially beneficial for link prediction tasks.

\begin{table*}[]
\caption{Experimental results for link prediction. The best results are highlighted in bold, the second-best results are underlined.}
\label{tab:lp}
\begin{threeparttable}
\begin{tabular}{@{}c|cccccccc@{}}
\toprule
Metrics               & Method   & Cora                & Citeseer            & Pubmed              & Photo               & Computer              & CS                  & Physics             \\ \midrule
\multirow{10}{*}{AP}  & DGI      & 93.60±1.14          & 96.18±0.68          & 95.65±0.26          & 81.01±0.47          & 82.05±0.50            & 92.79±0.31          & 92.10±0.29          \\
                      & MVGRL    & 92.95±0.82          & 89.37±4.55          & 95.53±0.30          & 63.43±2.02          & 91.73±0.40            & 89.14±0.93          & -                   \\
                      & GRACE    & 82.36 0.24          & 86.92±1.11          & 93.26±1.20          & 81.18 ± 0.37        & 83.12±0.23            & 83.90±2.20          & 82.20±1.06          \\
                      & GCA      & 80.87±4.11          & 81.93±1.76          & 93.31±0.75          & 65.17±10.11         & 89.50±0.64            & 83.24±1.16          & 82.80±4.46          \\
                      & CCA-SSG  & 93.74±1.15          & 95.06 ± 0.91        & 95.97±0.23          & 67.99±1.60          & 69.47±1.94            & 96.40±0.30          & 96.26±0.10          \\
                      & CAN      & 94.49±0.60          & 95.49±0.61          & -                   & 96.68±0.30          & 95.96±0.38            & -                   & -                   \\
                      & SIG-VAE  & 94.79±0.71          & 94.21±0.53          & 85.02±0.49          & 94.53±0.93          & 91.23±1.04            & 94.93±0.37          & 98.85±0.12          \\
                      & GraphMAE & 89.52±0.01          & 74.50±0.04          & 87.92±0.01          & 77.18 0.02          & 75.80±0.01            & 83.58±0.01          & 86.44±0.03          \\
                      & SeeGERA  & 95.92±0.68          & 97.33±0.46          & 97.87±0.20          & \textbf{98.48±0.06} & \textbf{97.50 ± 0.15} & 98.53±0.18          & \textbf{99.18±0.04} \\ \cmidrule(l){2-9} 
                      & Ours     & \textbf{98.03±0.10} & \textbf{97.76±0.38} & \textbf{98.86±0.04} & 97.47±0.20          & 97.25±0.04            & \textbf{99.06±0.02} & 99.01±0.003         \\ \midrule
\multirow{11}{*}{AUC} & DGI      & 93.88±1.00          & 95.98±0.72          & 96.30±0.20          & 80.95±0.39          & 81.27±0.51            & 93.81±0.20          & 93.51±0.22          \\
                      & MVGRL    & 93.33±0.68          & 88.66±5.27          & 95.89±0.22          & 69.58±2.04          & 92.37±0.78            & 91.45±0.67          & -                   \\
                      & GRACE    & 82.67±0.27          & 87.74±0.96          & 94.09±0.92          & 81.72±0.31          & 82.94±0.20            & 85.26±2.07          & 83.48±0.96          \\
                      & GCA      & 81.46±4.86          & 84.81±1.25          & 94.20±0.59          & 70.02±9.66          & 89.92±0.91            & 84.35±1.13          & 85.24±5.41          \\
                      & CCA-SSG  & 93.88±0.95          & 94.69±0.95          & 96.63±0.15          & 73.98±1.31          & 75.91±1.50            & 96.80±0.16          & 96.74±0.05          \\
                      & CAN      & 93.67±0.62          & 94.56±0.68          & -                   & 97.00±0.28          & 96.03±0.37            & -                   & -                   \\
                      & SIG-VAE  & 94.10±0.68          & 92.88±0.74          & 85.89±0.54          & 94.98±0.86          & 91.14±1.10            & 95.26±0.36          & 98.76±0.23          \\
                      & GraphMAE & 90.70±0.01          & 70.55±0.05          & 69.12±0.01          & 77.42±0.02          & 75.14±0.02            & 91.47±0.01          & 87.61±0.02          \\
                      & S2GAE    & 93.52±0.23          & 93.29±0.49          & 98.45±0.03          & -                   & -                     & -                   & -                   \\
                      & SeeGERA  & 95.50 ± 0.71        & 97.04±0.47          & 97.87±0.20          & \textbf{98.64±0.05} & \textbf{97.70 ±0.19}  & 98.42±0.13          & \textbf{99.03±0.05} \\ \cmidrule(l){2-9} 
                      & Ours     & \textbf{97.93±0.10} & \textbf{97.72±0.39} & \textbf{98.76±0.05} & 97.70±0.20          & 97.35±0.03            & \textbf{99.25±0.02} & \textbf{99.03±0.04} \\ \bottomrule
\end{tabular}
\begin{tablenotes}
        \footnotesize
        \item[] \scriptsize "-" indicates that unavailable code or out-of-memory.
    \end{tablenotes}
\end{threeparttable}
\end{table*}

\vpara{Graph Classification.}
\begin{table*}[htb]
\caption{Graph classification results on 7 datasets. The best results are highlighted in bold, the second-best results are underlined.}
\label{tab:gc}
\begin{tabular}{@{}ccccccccc@{}}
\toprule
\multicolumn{1}{l}{}           & Dataset   & \multicolumn{1}{l}{IMDB\_B} & \multicolumn{1}{l}{IMDB\_M} & \multicolumn{1}{l}{PROTEINS} & \multicolumn{1}{l}{COLLAB} & \multicolumn{1}{l}{MUTAG} & \multicolumn{1}{l}{REDDIT-B} & NCI \\ \midrule
\multirow{2}{*}{Graph Kernels} & WL        & 72.30±3.44                  & 46.95±0.46                       & 72.92±0.56                   & -                               & 80.72±3.00                & 68.82±0.41                   & 80.31±0.46              \\
                               & DGK       & 66.96±0.56                  & 44.55±0.52                       & 73.30±0.82                   & -                               & 87.44±2.72                & 78.04±0.39                   & 80.31±0.46              \\ \midrule
\multirow{2}{*}{Unsupervised}  & graph2vec & 71.10±0.54                  & 50.44±0.87                       & 73.30±2.05                   & -                               & 83.15±9.25                & 75.78±1.03                   & 80.31±0.46              \\
                               & Infograph & 73.03±0.87                  & 49.69±0.53                       & 74.44±0.31                   & 70.65±1.13                      & 89.01±1.13                & 82.50±1.42                   & 76.20±1.06              \\ \midrule
\multirow{5}{*}{Contrastive}   & GraphCL   & 71.14±0.44                  & 48.58±0.67                       & 74.39±0.45                   & 71.36±1.15                      & 86.80±1.34                & \underline{89.53±0.84}             & 77.87±0.41              \\
                               & JOAO      & 70.21±3.08                  & 49.20±0.77                       & 74.55±0.41                   & 69.50±0.36                      & 87.35±1.02                & 85.29±1.35                   & 78.07±0.47              \\
                               & GCC       & 72.0                        & 49.4                             & -                            & 78.9                            & -                         & \textbf{89.8}                & -                       \\
                               & MVGRL     & 74.20±0.70                  & 51.20±0.50                       & -                            & -                               & 89.70±1.10                & 84.50±0.60                   & -                       \\
                               & InfoGCL   & 75.10±0.90                  & 51.40±0.80                       & -                            & 80.00±1.30                      & \textbf{91.20±1.30}       & -                            & 80.20±0.60              \\ \midrule
\multirow{3}{*}{Generative}    & GraphMAE  & 75.52±0.66                  & 51.63±0.52                       & 75.30±0.39                   & 80.32±0.46                      & 88.19±1.26                & 88.01±0.19                   & 80.40±0.30              \\
                               & S2GAE     & \underline{75.76±0.62}          & \underline{51.79±0.36}               & \textbf{76.37±0.43}        & \underline{81.02±0.53}              & 88.26±0.76              & 87.83±0.27                 & \underline{80.80±0.24}   \\
                               & Ours & \textbf{76.27±0.37}         & \textbf{51.96±0.53}              & \underline{75.41±0.33}             & \textbf{81.92±0.01}           & \underline{89.36±0.00}        & 89.02±0.02                 & \textbf{80.93±0.02}      \\ \bottomrule
\end{tabular}
\end{table*}
We further extended our evaluation to include graph classification tasks and report the results in Table~\ref{tab:gc}. \model exhibited strong performance across all datasets, yielding either the best or comparable results against the tested baselines. These outcomes validate the effectiveness of our method in tackling not only node- and edge-level tasks but also graph-level tasks such as graph classification. The consistently high performance of our method across different datasets and in comparison to various baselines underscores the robustness and adaptability of our generative-based unified framework. This demonstrates the method's universal applicability for efficiently capturing and representing the unique characteristics and structures of disparate graph data, offering a compelling solution for graph classification tasks.

\vpara{Transfer Learning}
\begin{table*}[htb]
\caption{Transfer learning results on 8 datasets. The best results are highlighted in bold, the second-best results are underlined.}
\label{tab:transfer}

\begin{tabular}{@{}llllllllll@{}}
\toprule
            & BBBP              & Tox21             & ToxCast           & SIDER             & ClinTox           & MUV               & HIV               & BACE              & Avg.          \\ \midrule
ContextPred & 64.3±2.8          & {\ul 75.7±0.7}    & 63.9±0.6          & 60.9±0.6          & 65.9±3.8          & 75.8±1.7          & 77.3±1.0          & 79.6±1.2          & 70.4          \\
AttrMasking & 64.3±2.8          & \textbf{76.7±0.4} & {\ul 64.2±0.5}    & 61.0±0.7          & 71.8±4.1          & 74.7±1.4          & 77.2±1.1           & 79.3±1.6          & 71.1          \\
Infomax     & 68.8±0.8          & 75.3±0.5          & 62.7±0.4          & 58.4 ±0.8         & 69.9±3.0          & 75.3±2.5          & 76.0±0.7          & 75.9±1.6          & 70.3          \\
GraphCL     & 69.7±0.7          & 73.9±0.7          & 62.4±0.6          & 60.5±0.9          & 76.0±2.7          & 69.8±2.7          & {\ul 78.5±1.2}          & 75.4±1.4          & 70.8          \\
JOAO        & 70.2±1.0          & 75.0±0.3          & 62.9±0.5          & 60.0±0.8          & 81.3±2.5          & 71.7±1.4          & 76.7±1.2          & 77.3±0.5          & 71.9          \\
GraphLoG    & \textbf{72.5±0.8} & {\ul 75.7±0.5}    & 63.5±0.7          & {\ul 61.2±1.1}    & 76.7±3.3          & {\ul 76.0±1.1}    &  77.8±0.8    & \textbf{83.5±1.2} & 73.4          \\
GraphMAE    & 72.0±0.6          & 75.5±0.6          & 64.1±0.3          & 60.3±1.1          & {\ul 82.3±1.2}    & \textbf{76.3±2.4} & 77.2±1.0          & {\ul 83.1±0.9}    & {\ul 73.8}    \\ \midrule
Ours        & {\ul 72.4±0.5}    & 74.9±0.7          & \textbf{64.4±0.8} & \textbf{64.0±0.4} & \textbf{83.9±0.2} & 75.5±2.8          & \textbf{79.8±0.6} & 82.0±0.7          & \textbf{74.6} \\ \bottomrule
\end{tabular}
\end{table*}
Lastly, we conducted a series of transfer learning experiments to assess the transferability of \model. Our experiments aimed to evaluate whether the powerful representations learned by \model could be generalized to new, unseen tasks. Specifically, we pre-trained our model on a corpus of two million unlabeled molecular samples from the ZINC15~\cite{zinc} dataset. We then fine-tuned and tested our approach on eight graph classification datasets. The datasets were divided in a scaffold-split manner to simulate realistic application scenarios, aligning with common practices in cheminformatics. The results of these experiments are presented in Table~\ref{tab:transfer}. \model achieved best result on 4 out of 8 datasets. Meanwhile, the performance on the remaining datasets is on par with the competitive baselines. Moreover, when evaluating the average scores across all eight datasets, our method outshone the others, achieving the most favorable overall results. Such outcomes underscore the excellent and stable transfer learning capabilities of our method, further portraying its robustness and the quality of the graph representations it generates.
\subsection{Ablation Study ($\mathcal{RQ}2$)}~\label{ablation_study}
\begin{figure}[htb]
    \centering
    \includegraphics[width=1.0\linewidth]{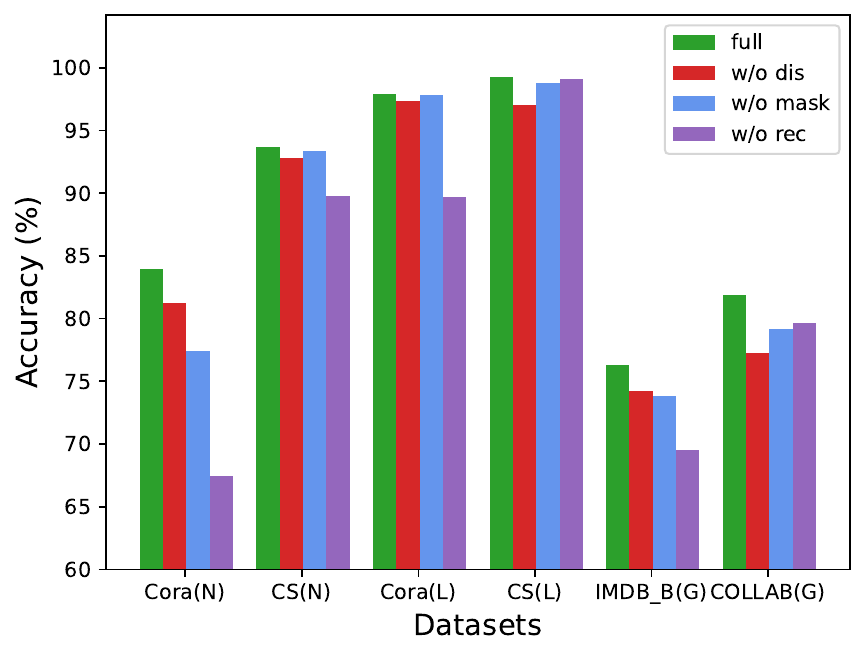}    
    \caption{Ablation studies of main components.}
    \label{fig:ablation}
\end{figure}
To validate the impact of various modules of \model, we conduct ablation experiments by removing certain module. We use "full" to denote the intact \model. "w/o dis" signify the model without the discriminator module. "w/o rec" means \model abandon the reconstruction task, rely only on the discriminator to optimize. Lastly, “w/o mask" represent the model without the mask module, where the mask rate is set to zero. We select two datasets each from node-, edge-, and graph-level tasks to conduct the ablation studies. As can be inferred from Figure~\ref{fig:ablation}, all the operative modules make a positive contribution to the overall performance of the model. Removal of any of these modules resulted in a drop in model performance, substantiating their importance in correct and efficient graph representation learning. 
\subsection{Parameter Analysis ($\mathcal{RQ}3$)}
 \begin{figure}[htb]
     \centering
     \includegraphics[width=1.0\linewidth]{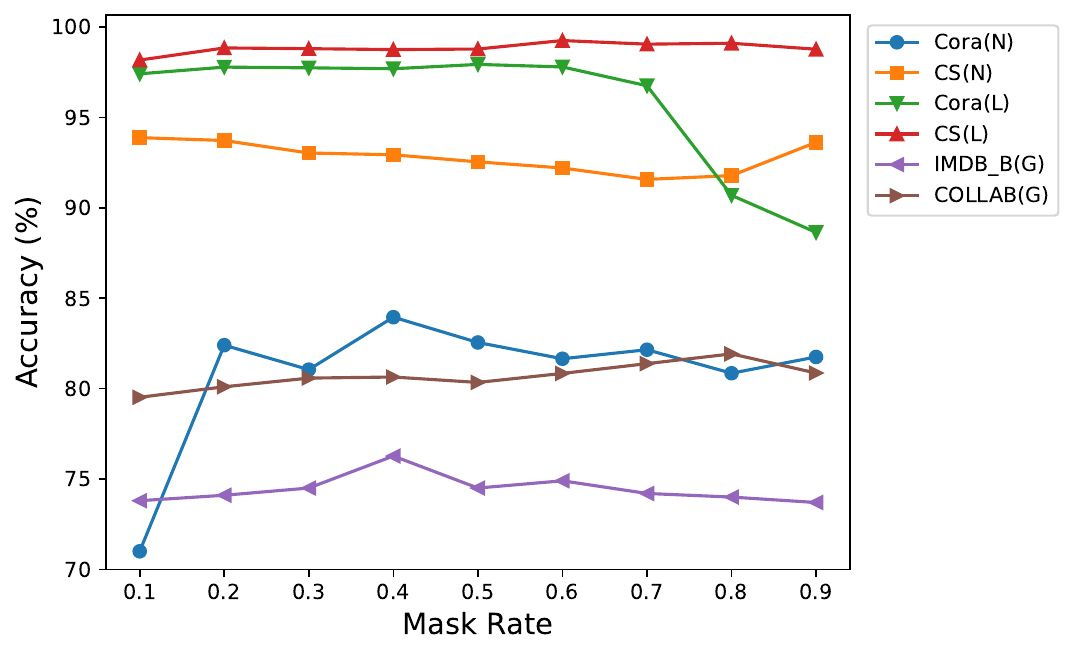}
     \caption{Impact of mask ratio.}
     \label{fig:mask}
 \end{figure}
\vpara{Mask Rate.} In the previous ablation study, we confirmed the positive contributions of the mask ratio, thus we further exam \model's sensitivity by setting the mask ratio between 0.1 to 0.9 with step of 0.1. We select two datasets each from the tasks of node classification, link prediction, and graph classification datasets. 

As can be seen from Figure~\ref{fig:mask}, the mask ratio has distinct impact on different graph tasks. The node- and edge-level are more sensitive to the mask ratio compared to the graph tasks. Beside, the optimal mask ratio for different graph tasks is different. Generally, too large or small mask ratio exhibit negative impact. Besides, the result suggests that \model is relatively stable for mask ratio between 0.2 to 0.7.

\section{CONCLUSION}
In this paper, we propose an adversarially masked autoencoder, \model, that addresses multi-task graph learning. \model accepts arbitrary graphs as input and reformulates them into subgraphs, then using a masked GAE to reconstruct the input subgraph and generate the reconstructed graph. Besides, \model adopt a GNN Readout as a discriminator to determine the authenticity between the reformulated graph and the reconstructed graph. The design of \model not only ensures its ability in process various kinds of graphs, but also guarantee the robustness across different domains. \model exhibits strong performance across various graph tasks, demonstrates its effectiveness and universality.

\bibliographystyle{ACM-Reference-Format}
\clearpage
\bibliography{sample-base}

\clearpage
\appendix
\section{Motivation}
\subsection{The "Pre-training + Fine-tuning" paradigm}
\begin{figure}[h]
    \centering
    \includegraphics[width=0.8\linewidth]{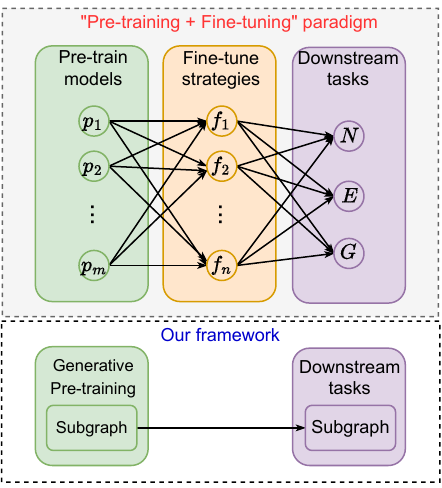}
    \caption{Comparison of our approach with existing ``pre-training + fine-tuning'' paradigm}
    \label{fig:pretrain}
\end{figure}

\section{Methodology supplement}
\subsection{Random Walk}~\label{app:random_walk}
\begin{algorithm}[h]
    \caption{Random walk with restart(RWR) algorithm}\label{alg:rwr}    
    \KwIn{Input graph $\gG$, target node $v$, restart probability $p$, expected number of collected nodes $RWR(v)$. \\ }
    \KwOut{A sampled node set $\gS$;}
    $\gS \leftarrow \{\}, u \leftarrow v$; \\
    \While{$\text{len}(\gS) < RWR(u)$}{
	    \If{$u$ not in $\gS$}{
	        $\gS$.append($u$)
	    }
	    Sample $r$ from Uniform distribution $U(0,1)$; \\
	    \If{$r< p$}{
	        $u \leftarrow v$;
	    }
	    \Else{
	        {Randomly sample $\hat{u}$ from $u$'s neighbors}; \\
	        {$u \leftarrow \hat{u}$};
	    }
	}
    \Return{$\gS$}
\end{algorithm}

\begin{figure}[t!]
  \centering
  \subfigure[RWR to construct node subgraph.]{
		\label{rwr_node}
		\includegraphics[width=0.23\textwidth]{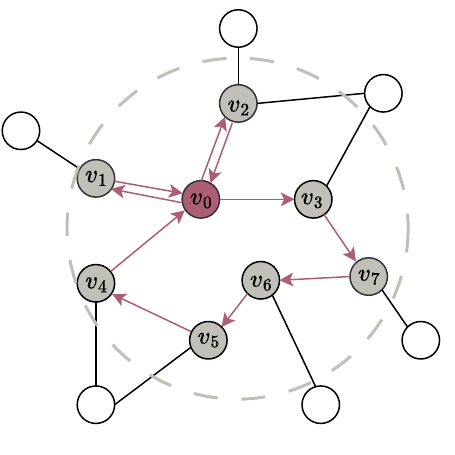}}
    \subfigure[RWR to construct edge subgraph.]{
		\label{rwr_edge}
		\includegraphics[width=0.23\textwidth]{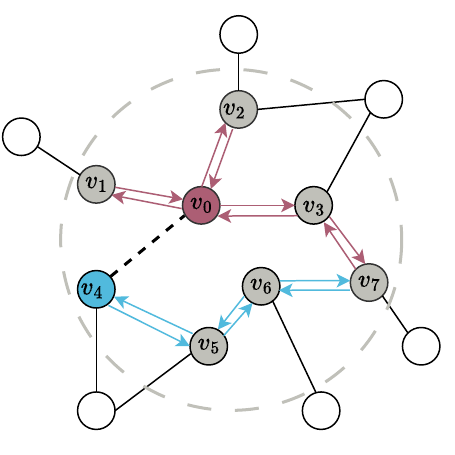}}		
    \caption{An illustration of RWR to construct subgraph for node and edge-level subgraphs.} 
  \setlength{\abovecaptionskip}{0cm}
\end{figure}

\subsection{Reconstruction criterion}~\label{reconstruction_criterion}
The scaled cosine error is formulated as:
\begin{equation}
\label{eq:sce}
\mathcal{L}_{\mathrm{sce}}=\frac{1}{|{\mathcal{V}}|} \sum_{v_i \in {\mathcal{V}}}\left(1-\frac{x_i^T \bar{x}_i}{\left\|x_i\right\| \cdot\left\| \bar{x}_i \right\|} \right)^\gamma, \gamma \geq 1,
\end{equation}
where $x_i \in \mathbf{X}$ stands for the node features of the original graph $\mathcal{G}$, $\bar{x}_i  \in \mathbf{\bar{X}}$ denotes the node features of the reconstructed graph $\mathcal{\bar{G}}$ generated by the decoder. The scaling factor $\gamma$ is a hyperparameter. When $\gamma >1$, we can weight down the easy samples in training with a $\gamma$ power of the cosine error. For high-confidence predictions, the corresponding cosine error is usually less than 1 and decays faster with a scaling factor $\gamma>1$.

\subsection{Background}~\label{app:background}
\vpara{Graph Neural Network (GNN).}
GNNs have demonstrated strong capabilities in graph representation and tackling graph tasks~\cite{cai2005mining, jin2021application, shi2016survey}. In this paper, we employ GNNs for graph representation. Specifically, for a single node $v \in \mathcal{V}$ on the graph, it aggregates the representations of its neighboring nodes as:
\begin{equation}
    e_v^{l} = AGG(e_v^{l-1}, {e_u^{l-1} \mid u\in{\mathcal{N}_v}}), 
\end{equation}
where $N_v$ represents the neighbors of $v$ and $e_v$ denotes the representation of $v$ at the $l$-th layer. For a subgraph $sg$, we utilize simple readout modules, e.g., max or mean pooling, on $sg$ to represent the subgraph representation as:
\begin{equation}
E_{sg} = readout(\left [ e_i, e_{i+1},...,e_{i+n} \right ] ).
\end{equation}

\vpara{Masked Graph Autoencoder.}
The Masked Graph Autoencoder (GAE) is a self-supervised model typically comprising an encoder $f_E$ and a decoder $f_D$. Given an input graph $\mathcal{G}$, we randomly sample a fixed proportion of node indices $M$. $\mathcal{G}$ can be divided into a masked graph $\mathcal{G}_m$ and an unmasked graph $\mathcal{G}_u$, with the corresponding feature matrix divided into $X_m = \{x^k \in M\}$ and $X_u = \{x^k \notin M\}$. During training, the encoder $f_E$ first maps the corrupted graph to a hidden representation $\mathcal{H} \in \mathbb{R}^{N \times d_h}$. Subsequently, the decoder $f_D$ attempts to reconstruct the feature or adjacency matrix using the hidden representation $\mathcal{H}$. Conventionally, the mean-squared error (MSE) or scaled cosine error (SCE) is employed as the reconstruction criterion.

\vpara{Generative Adversarial Network (GAN).}
Generative Adversarial Networks (GANs)~\cite{gan_survey} are robust networks that excel at generating high-quality data. GANs typically consist of a generator network ($G$) that produces synthetic samples closely resembling real data, and a discriminator network ($D$) to estimate the authenticity of a given sample. The generator $G$ strives to deceive the discriminator, while the discriminator model $D$ works to avoid being tricked. Collectively, $G$ and $D$ participate in a minimax game as follows:
\begin{equation}
\begin{aligned}
\min _G \max _D L(D, G) &= \mathbb{E}_{x \sim p_r(x)}[\log D(x)] \\
        &+ \mathbb{E}_{x \sim p_g(x)}[\log (1-D(x))],
\end{aligned}
\end{equation}
where $p_r(x)$ represents the distribution of real data over real samples $x$, and $p_g(x)$ denotes the generator's distribution over $x$.

\subsection{Training and Inference}~\label{training_inference}
\vpara{Training.} The training process is identical for node-, edge-, and graph-level tasks. The training process accepts unified meta-structure as input, these subgraphs are subsequently fed into the masked GAE for training. More specifically, we begin by randomly masking the node attribute of the graph, then feeding them to the GAE encoder. The graph is reconstructed by the GAE decoder, with node features forming the target for generative pre-training. The reconstructed graph from the masked GAE, along with the original graph, is subsequently sent to the discriminator, which determines whether the input graph has been generated by the GAE. 
As depicted in Algorithm~\ref{alg:procedure}, the generator and the discriminator are iteratively optimized during a single training epoch. The masked GAE (generator) initially reconstructs the graph as well as generates the output graph based on $\L_{sce}$ and $\L_{gen}$, and further optimizes them together. The GNN readout module (discriminator) then discerns the authenticity of the input and reconstructed graphs.

\vpara{Inference.} The inference stage only involves the encoder, which also accepts subgraph as input, thereby ensuring consistency between the pre-training and inference stages. Assuming the encoder accepts $B$ subgraphs for inference, it maps the subgraph's feature into a dense representation as $\mathbf{h}=f_E(\gG), \mathbf{h} \in \mathbb{R}^{N \times d_h}$, where $d_h$ represents the hidden dimension. Subsequently, a non-parametric graph pooling readout module (max, mean, or summation pooling) aggregates the graph-level representation into $\mathbf{z} \in \mathbb{R}^{N \times 1}$. Finally, the label of each $sg$ is ascertained by the $Softmax$ function. Similar to the training stage, the inference process is task-agnostic, i.e., identical for different graph tasks.

\section{Additional Experiment Information}

\subsection{Datasets}~\label{app:data}
\begin{table}[htb]
\centering
\caption{Statistical information on datasets for node classification and link prediction.}
\label{tab:data_nl}
\begin{tabular}{@{}lllll@{}}
\toprule
Datasets                              & Nodes & Edges  & Features & Classes \\ \midrule
\multicolumn{1}{l|}{Cora}             & 2708  & 5278   & 1433     & 7       \\
\multicolumn{1}{l|}{Citeseer}         & 3327  & 4676   & 3703     & 6       \\
\multicolumn{1}{l|}{Pubmed}           & 19717 & 88651  & 500      & 3       \\
\multicolumn{1}{l|}{Coauthor CS}      & 18333 & 327576 & 6805     & 15      \\
\multicolumn{1}{l|}{Coauthor Physics} & 34493 & 991848 & 8451     & 5       \\
\multicolumn{1}{l|}{Amazon Computer}  & 13752 & 574418 & 767      & 10      \\
\multicolumn{1}{l|}{Amazon Photo}     & 7650  & 287326 & 745      & 8       \\ \bottomrule
\end{tabular}
\end{table}
To comprehensively evaluate \model on node- and edge-level tasks, we conduct experiments on seven public datasets across various categories, including citation networks (Cora, Citeseer, and Pubmed), co-authorship graphs (Coauthor CS and Coauthor Physics), and co-purchase graphs (Amazon Computer and Amazon Photo).

For graph classification tasks, we perform experiments on an additional set of seven public datasets, which include MUTAG, IMDB-B, IMDB-M, PROTEINS, COLLAB, REDDITB, and NCI1~\cite{10.1145/2783258.2783417}. Among these, PROTEINS, MUTAG, and NCI1 are bioinformatics datasets, while the rest are social network datasets. Each dataset within this collection comprises a set of graphs with corresponding labels.  

For the transfer learning task, we test \model with molecular property prediction. \model is initially pre-trained on 2 million unlabeled molecules based on ZINC15~\cite{zinc}, and subsequently fine-tuned on 8 classification benchmark datasets contained in MoleculeNet~\cite{moleculenet}.
\begin{table*}[htbp]
    \centering    
    \renewcommand\arraystretch{1.05}
    \caption{Statistics of datasets for molecular property prediction. ``ZINC'' is used for pre-training.}
    \begin{tabular}{@{}c|ccccccccc}  
        \toprule[1.2pt]
                      & ZINC & BBBP & Tox21 & ToxCast & SIDER & ClinTox & MUV & HIV & BACE \\
        \midrule
        \# graphs      & 2,000,000 & 2,039 & 7,831 & 8,576 & 1,427 & 1,477 & 93,087 & 41,127 & 1,513\\     
        \# binary prediction tasks     & - & 1 & 12 & 617 & 27 & 2 & 17 & 1 & 1   \\            
        Avg. \# nodes  & 26.6 & 24.1 & 18.6 & 18.8 &  33.6 & 26.2 & 24.2 & 24.5 & 34.1 \\ 
        \bottomrule[1.2pt]
    \end{tabular}
    \label{tab:stat_mol}
\end{table*}

We strictly follow the same experimental procedures as previous works~\cite{GraphMAE, graphmae2, SeeGera}, including data splitting, evaluation protocols, metrics, etc.

\subsection{Baselines}~\label{app:baselines}
For node classification tasks, we compare our method with 10 state-of-the-art (SOTA) baselines divided into two categories: (1) Supervised methods, which include GCN~\cite{gcn} and GAT~\cite{gat}. (2) Self-supervised methods, consisting of contrastive learning approaches such as DGI~\cite{veličković2018deep}, MVGRL~\cite{hassani2020contrastive}, GRACE~\cite{zhu2020deep}, and CCA-SSG~\cite{zhang2021canonical}, as well as generative methods including VGAE~\cite{kipf2016semi}, GraphMAE~\cite{hou2022graphmae}, GraphMAE2~\cite{graphmae2}, and SeeGera~\cite{li2023seegera}.

For link prediction tasks, we compare our method with 10 SOTA competitors, comprising 5 contrastive self-supervised learning methods: DGI~\cite{dgi}, MVGRL~\cite{mvgrl}, GRACE~\cite{grace}, GCA~\cite{gca}, CCA-SSG~\cite{cca-ssg} and 5 generative self-supervised learning methods: CAN~\cite{can}, SIG-VAE~\cite{sig-vae}, GraphMAE~\cite{GraphMAE}, S2GAE~\cite{S2GAE}, and SeeGERA~\cite{li2023seegera}.

For graph classification tasks, we compare \model with 11 SOTA baselines of three categories: (1) Graph kernel methods, which include the WeisfeilerLehman sub-tree kernel (WL)~\cite{shervashidze2011weisfeiler} and Deep Graph Kernel (DGK)~\cite{yanardag2015deep}. (2) Unsupervised methods, including graph2vec and Infograph~\cite{sun2019infograph}. (3) Self-supervised methods, which consist of GraphCL~\cite{you2020graph}, JOAO~\cite{you2021graph}, GCC~\cite{qiu2020gcc}, MVGRL, InfoGCL~\cite{xu2021infogcl}, GraphMAE, and S2GAE~\cite{S2GAE}. 

For transfer learning tasks, we compare \model with 7 baselines, including Infomax~\cite{infomax}, AttrMasking, ContextPred, and SOTA self-supervised learning methods such as GraphCL~\cite{graphcl}, JOAO~\cite{joao}, GraphLoG~\cite{GraphLoG}, and GraphMAE~\cite{GraphMAE}.

\subsection{Experimental Settings}~\label{app:experient_settings}
For all baselines, if official code implementations are available, we reproduce their experimental results while adhering to consistent experimental settings. If the original papers provide results under consistent settings, we directly incorporate their provided results. For \model, we employ GCN or GAT as the backbone for both the generator and discriminator. For node and graph classification, we report Accuracy (Acc) as the evaluation metric. For link prediction, we utilize the Area Under the Curve (AUC) and Average Precision (AP) as evaluation metrics. All experiments are conducted on a single V100 32GB GPU, with each experiment repeated five times. We report the mean and standard deviation of the results.

\begin{table}[htb]
\centering
\caption{Statistical information on datasets for graph classification}
\label{tab:data_g}
\begin{tabular}{@{}llll@{}}
\toprule
Datasets                      & Graphs & Avg Nodes & Classes \\ \midrule
\multicolumn{1}{l|}{IMDB-B}   & 1000   & 19.77     & 2       \\
\multicolumn{1}{l|}{IMDB-M}   & 1500   & 13        & 3       \\
\multicolumn{1}{l|}{PROTEINS} & 1113   & 39.1      & 2       \\
\multicolumn{1}{l|}{COLLAB}   & 5000   & 74.49     & 3       \\
\multicolumn{1}{l|}{MUTAG}    & 188    & 17.9      & 2       \\
\multicolumn{1}{l|}{REDDIT-B} & 2000   & 429.61    & 2       \\
\multicolumn{1}{l|}{NCI1}     & 4110   & 29.8      & 2       \\ \bottomrule
\end{tabular}
\end{table}
\end{document}

%% file: math_commands.tex
\usepackage{amsmath,amsfonts,bm}

\def\eqref#1{equation~\ref{#1}}

\def\1{\bm{1}}

\DeclareMathAlphabet{\mathsfit}{\encodingdefault}{\sfdefault}{m}{sl}
\SetMathAlphabet{\mathsfit}{bold}{\encodingdefault}{\sfdefault}{bx}{n}

\def\gG{{\mathcal{G}}}

\def\gL{{\mathcal{L}}}

\def\gS{{\mathcal{S}}}
\def\gT{{\mathcal{T}}}

\def\gV{{\mathcal{V}}}

